\documentclass[letterpaper, 10 pt, conference]{ieeeconf}
\pdfoutput=1
\IEEEoverridecommandlockouts                
\usepackage{graphicx}
\usepackage{float}
\usepackage{color}
\usepackage{array}
\usepackage{multirow}
\usepackage{multicol}
\usepackage{booktabs}
\usepackage{rotating}
\usepackage{csquotes}
\let\labelindent\relax  
\usepackage{enumitem}
\usepackage{textcomp}  
\usepackage{gensymb}
\usepackage[pagebackref=true,breaklinks=true,colorlinks,bookmarks=false]{hyperref}

\newcommand{\tabref}[1]{Table~\ref{#1}}
\newcommand{\figref}[1]{Fig.~\ref{#1}}
\newcommand{\myparagraph}[1]{\vspace{0.03in}\noindent\textbf{#1}}

\newcommand{\eg}{\textit{e}.\textit{g}. }

\makeatletter
\let\NAT@parse\undefined
\makeatother
\usepackage[numbers]{natbib}

\title{\LARGE \bf ClearGrasp: \\3D Shape Estimation of Transparent Objects for Manipulation \vspace{-3mm}}

\author{Shreeyak S. Sajjan$^{1}$  \quad　Matthew Moore$^{1}$  \quad Mike Pan$^{1}$  \quad Ganesh Nagaraja$^{1}$  \\ Johnny Lee$^{2}$ \quad  Andy Zeng$^{2}$  \quad  Shuran Song$^{2,3}$
\vspace{0.1cm} \\ 
$^{1}$ Synthesis.ai \quad\quad
$^{2}$ Google \quad\quad
$^{3}$ Columbia University \quad\quad
}

\begin{document}

\maketitle
\thispagestyle{empty}
\pagestyle{empty}

\begin{abstract}
Transparent objects are a common part of everyday life, yet they possess unique visual properties that make them incredibly difficult for standard 3D sensors to produce accurate depth estimates for. In many cases, they often appear as noisy or distorted approximations of the surfaces that lie behind them. To address these challenges, we present ClearGrasp -- a deep learning approach for estimating accurate 3D geometry of transparent objects from a single RGB-D image for robotic manipulation. Given a single RGB-D image of transparent objects, ClearGrasp uses deep convolutional networks to infer surface normals, masks of transparent surfaces, and occlusion boundaries. It then uses these outputs to refine the initial depth estimates for all transparent surfaces in the scene. To train and test ClearGrasp, we construct a large-scale synthetic dataset of over 50,000 RGB-D images, as well as a real-world test benchmark with 286 RGB-D images of transparent objects and their ground truth geometries. The experiments demonstrate that ClearGrasp is substantially better than monocular depth estimation baselines and is capable of generalizing to real-world images and novel objects. We also demonstrate that ClearGrasp can be applied out-of-the-box to improve grasping algorithms' performance on transparent objects. Code, data, and benchmarks will be released. Supplementary materials: \href{https://sites.google.com/view/cleargrasp}{https://sites.google.com/view/cleargrasp}\footnote{We would like to thank Ryan Hickman for managerial support, Ivan Krasin and Stefan Welker for fruitful technical discussions, Cameron (@camfoxmusic) for sharing 3D models of his potion bottles and Sharat Sajjan for helping on webpage design.
}

\end{abstract}

\section{Introduction}

Transparent objects are a common part of everyday life, from reading glasses to plastic bottles -- yet they possess unique visual properties that make them incredibly difficult for machines to perceive and manipulate. 
In particular, transparent materials (which are both refractive and specular) do not adhere to the geometric light path assumptions made in classic stereo vision algorithms. This makes it challenging for standard 3D sensors to produce accurate depth estimates for transparent objects, which often appear as noisy or distorted approximations of the surfaces that lie behind them. Hence, while considerable research has been devoted to robotic manipulation of objects using 3D data (\eg RGB-D images, point clouds) \cite{ten2018using,zeng2018robotic,mahler2017dex}, many of these algorithms cannot be immediately applied to transparent objects -- which remain critical for applications like dish washing or sorting/cleaning plastic containers.

\begin{figure}[t]
    \centering
    \includegraphics[width=\linewidth]{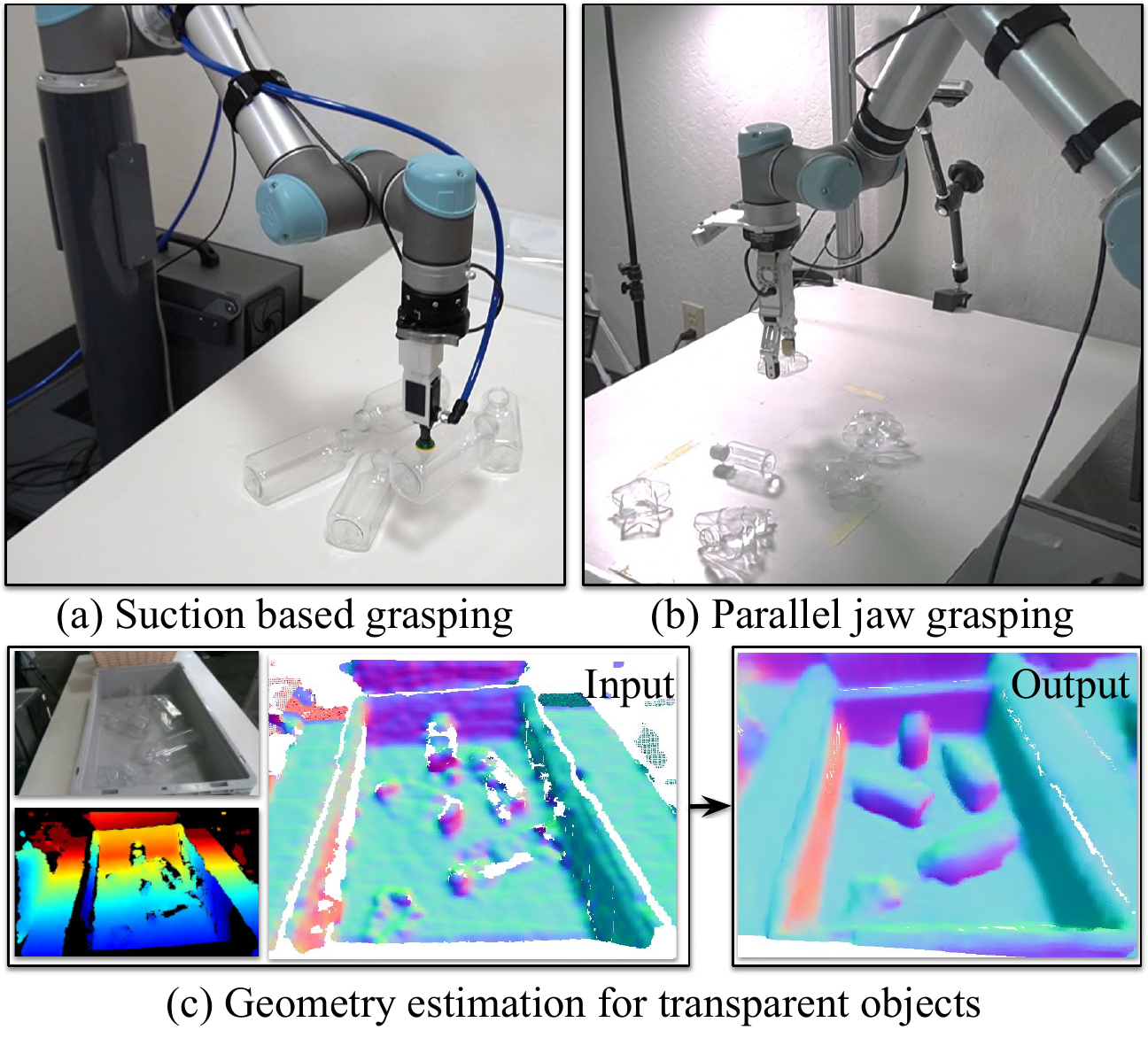}
\caption{\textbf{ClearGrasp} leverages deep learning with synthetic training data to infer accurate 3D geometry of transparent objects from a single RGB-D image. The estimated geometry can be directly used for downstream robotic manipulation tasks (\eg suction and parallel-jaw grasping).}
    \label{fig:robot_grasping}
    \vspace{-5mm}
\end{figure}

\begin{figure*}[t]
    \centering
    \includegraphics[width=0.9\linewidth]{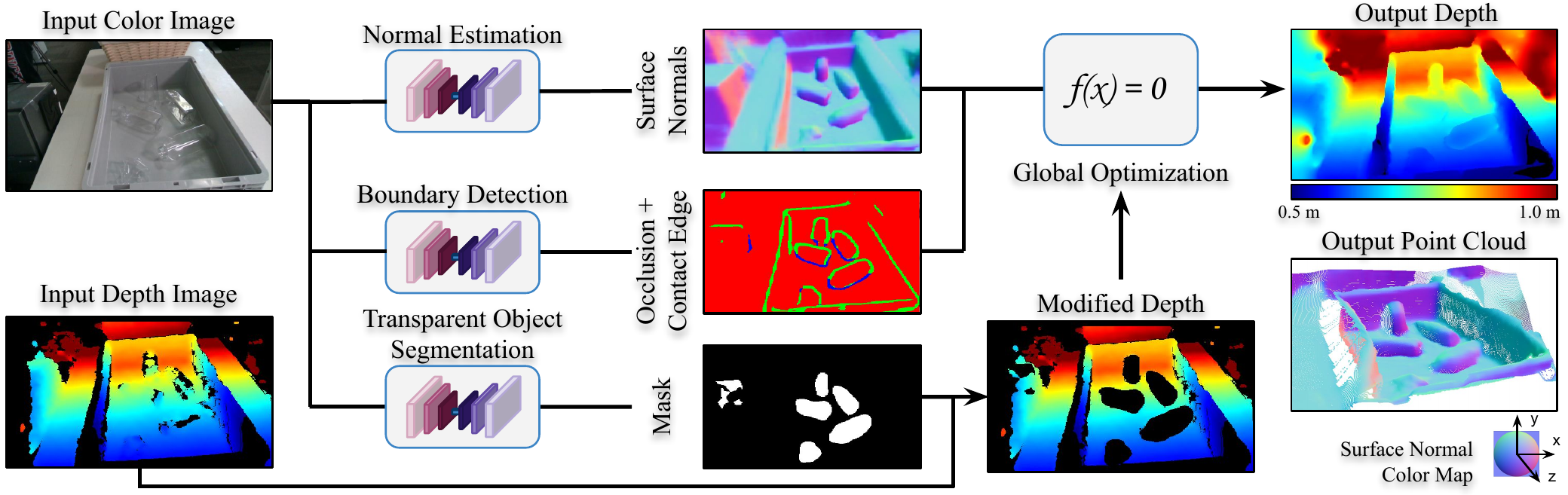}
    \caption{\label{fig:method} \textbf{Overview.} 
  Given an RGB-D image of a scene with transparent objects, ClearGrasp uses three networks to infer 1) surface normals, 2) masks of transparent surfaces, where depth is unreliable, and 3) occlusion and contact edges between the transparent surfaces and the rest of the scene. 
These outputs are then combined and used as input to a global optimization, which returns an adjusted depth map that corrects and completes the input depth. 
}
    \vspace{-4mm}
\end{figure*}

In this work, we present \textbf{ClearGrasp}, an algorithm that leverages deep learning with synthetic training data to infer accurate 3D geometry of transparent objects for robotic manipulation. The design of ClearGrasp is driven by the following three key ideas:
\begin{itemize}[leftmargin=*]
  \item Commodity RGB-D cameras often provide good depth estimates for typical non-transparent surfaces. Therefore, rather than directly estimating all geometry from scratch, we conjecture that correcting initial depth estimates from RGB-D cameras is more practical: enabling us to use the depth from the non-transparent surfaces to inform the depth of transparent surfaces. For this to work reliably, we propose to predict pixel-wise masks of transparent surfaces (to detect and remove unreliable depth), as well as occlusion and contact edges between transparent surfaces and the background (to extend reliable depth).

  \item The refractive and specular patterns appearing on transparent objects provide stronger visual cues for their curvature (\eg surface normals) than their absolute depth. This motivates using deep networks to infer surface normal information from RGB data, which we find to be substantially more reliable than directly inferring depth values.  
  
  \item While real-world ground truth 3D training data for transparent objects is difficult to obtain, we show that it is possible use high-quality rendered synthetic images with domain randomization as training data to obtain reasonable results on real-world data. Interestingly, we also find that by mixing synthetic training data with real-world out-of-domain data (\eg images without transparent objects), our model is able to generalize better to both real-world images and novel transparent objects unseen during training.
\end{itemize}

Our primary contributions are twofold. First, we propose an algorithm for estimating accurate 3D geometry of transparent objects from RGB-D images. 
Second, we construct a large-scale synthetic dataset of over 50,000 RGB-D images as well as a real-world test benchmark with 286 RGB-D images of transparent objects and their ground truth geometries. Our experiments demonstrate that ClearGrasp is capable of generalizing not only to transparent objects in the real-world, but also to novel objects unseen in training. ClearGrasp is substantially better than monocular depth estimation baselines, and our ablative studies show the importance of critical design decisions. We also demonstrate that ClearGrasp can be applied out-of-the-box with state-of-the-art manipulation algorithms to achieve 86\% and 72\% picking success rates with suction and parallel-jaw grasping respectively on a real-world robot platform. Code, data, pre-trained models, and benchmarks will be released.

\section{Related Work}

\myparagraph{Estimating geometry from color images.} Surface normal estimation is a popular problem tackled by deep convolutional networks \cite{wang2015surface, eigen2015predicting, zhang2016physically, Zeng_2019_CVPR}. While predicted surface normals are useful for tasks 
like shading \cite{Hudon_2018_ECCV_Workshops}, 2D-3D alignment \cite{bansal2016marr}, and face morphing \cite{kokkinosface}, it alone is insufficient to describe an object's complete 3D geometry, making it difficult to be directly used by manipulation algorithms that require 3D data (\eg depth images, point clouds) \cite{ten2018using,zeng2018robotic,mahler2017dex}. %
More recent works study how to obtain 3D data from color images by directly inferring depth images \cite{corr2014EigenPF,corr2016LainaRBTN,corr2016ChenFYD,ramamonjisoa2019sharpnet,hu2019revisiting,icra2019fastdepth,fu2018deep}, or filling in missing depth values in RGB-D images captured by commodity 3D cameras \cite{herrera2013depth, gong2013guided,zhang2018deepdepth}. 
However, none of these works explicitly handle transparent objects, for which ground truth 3D data is very difficult to obtain -- data from commodity 3D stereo cameras often have inaccurate or missing depth estimates for transparent surfaces.

\myparagraph{Recognizing transparent objects.}
Transparent objects have plagued computer vision since the inception of the field. Due to their refractive and reflective nature, their appearance can vary drastically according to background and illumination conditions. Classic methods for detecting transparent objects mostly relied on idiosyncrasies such as specular reflections or local characteristics of edges due to refraction \cite{phillips2011novel, fritz2009additive, mchenry2006geodesic, McHenry_2005_findingglass}. Later methods rely on deep learning models like SSD \cite{khaing2018transparent} or RCNN \cite{lai2015transparent} to predict bounding boxes enclosing transparent objects. \citet{seib2017friend} proposed a method to exploit sensor failures in depth images for transparent object localization using convolutional networks. \citet{wang2012glass} proposed localizing glass objects using a Markov Random Field to predict glass boundary and region jointly from multiple modalities from an RGB-D camera. Based on the localization, they recover depth readings by a piece-wise planar model. However, our method not only detects transparent objects, but also recovers detailed non-planar geometries, which are critical for manipulation algorithms.

\myparagraph{Estimating geometry of transparent objects.}
Works on estimating transparent object geometry are often studied in a constrained environment: 
For example, the work might assume a specific capturing procedure \cite{guo2019transparent, ji2017fusing, albrecht2013seeing}, known background pattern \cite{Qian_2016_CVPR, han2015fixed}, sensor type \cite{song2018depth} or known object 3D model \cite{phillips2016seeing, lysenkov2013pose, klank2011transparent}. \citet{lysenkov2013recognition} propose a method for the recognition and pose estimation of rigid transparent objects using a Kinect sensor. Using a segmentation mask of the transparent objects, 3D models of objects created at the training stage are fitted to extracted edges. Our approach is able to generalize to objects not seen during training and does not require prior knowledge of the 3D model of the objects or camera position. 

\myparagraph{Learning from synthetic data.}
Synthetic data has proven to be useful in various tasks such as depth estimation \cite{rematas2018soccer}, 3D semantic scene completion \cite{song2016ssc}, hand pose estimation \cite{deng2017handpose}, robotic grasping \cite{mahler2017dex}, automatic shading of sketches \cite{Hudon_2018_ECCV_Workshops}, and person re-identification for tracking \cite{Barbosa2018tracking}.
However, very few synthetic datasets support planar reflectors \cite{replica19arxiv}, let alone transparent objects. In our trials, we find that very high quality rendering and 3D models are required to synthesize representative imagery of transparent objects and their related artifacts \eg specular highlights and caustics.
Datasets that do contain transparent objects have been used to study refractive flow estimation \citet{chen2018tomnet}, semantic segmentation \cite{stets2019materialbased}, or relative depth \cite{stets2019single}.
These datasets are generated in a simplified setting (\eg rendered transparent objects in front of random images from COCO \cite{lin2014microsoft}). On the contrary, our method targets reconstructing detailed absolute depth of transparent objects within realistic environments.

\section{Method}
\label{sec:method}
Given a single RGB-D image of transparent objects, ClearGrasp first uses the color image as input to deep convolutional networks to infer a set of information: surface normals, masks of transparent surfaces, and occlusion boundaries. ClearGrasp then uses this information and the initial depth image as input to a global optimization, which outputs a new depth image that refines the initial depth estimates from the sensor for all transparent surfaces in the scene (Sec. \ref{sec:vision}).
For training and testing, we construct a synthetic dataset and a real-world benchmark for transparent objects (Sec. \ref{sec:data-syn}, \ref{sec:data-real}).
In Sec. \ref{sec:data-syn}, we demonstrate the application of ClearGrasp to a real-word robotic pick-and-place system.

\subsection{Estimating 3D Geometry of Transparent Object \label{sec:vision}}
\begin{figure}[t]
    \centering
    \includegraphics[width=\linewidth]{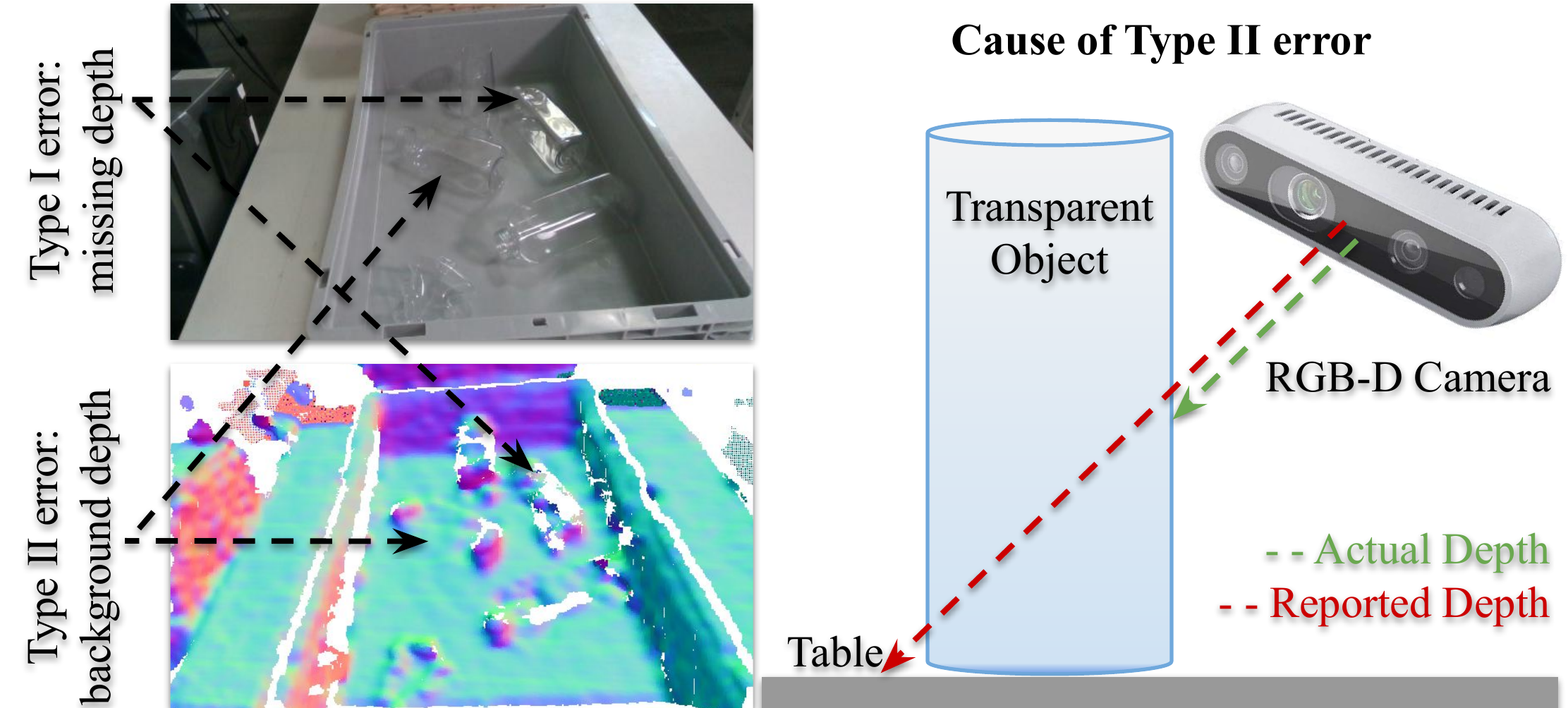}
    \caption{\textbf{Errors in depth for transparent objects:} Type I errors, missing depth, is often caused by specular reflections on the surface. Type II errors, inaccurate depth estimates (returns background depth instead of the object depth), is caused by transparency of the surface material. 
    }
    \label{fig:sensor_errors}
    \vspace{-4mm}
\end{figure}

\begin{figure}[ht]
\vspace{-1mm}
    \centering
    \includegraphics[width=1.0\linewidth]{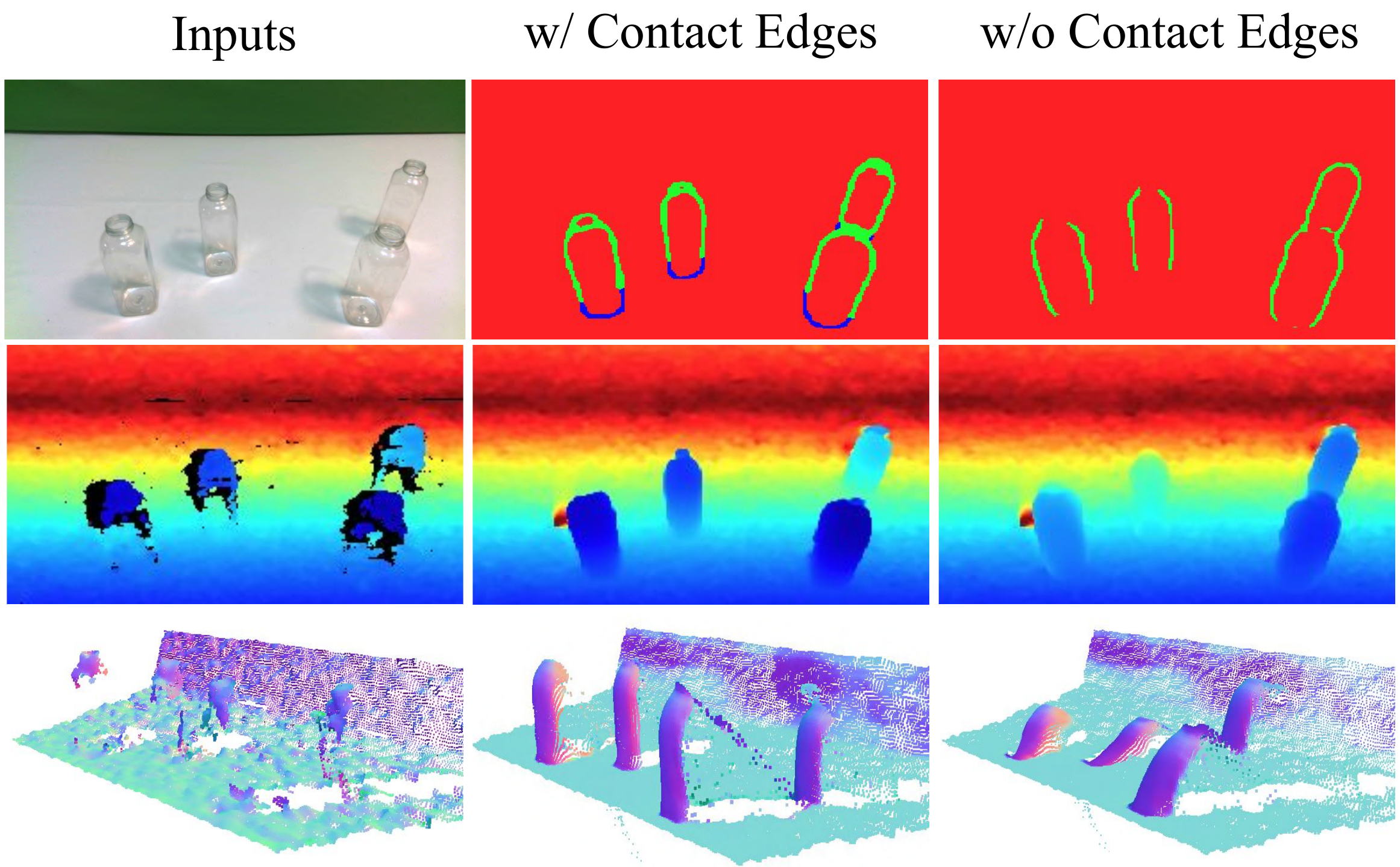}
    \caption{\textbf{Effects of contact edges. } By training our boundary estimation model with contact edges (middle column), ClearGrasp predicts better depth for transparent objects than without contact edges (right column).}
    \label{fig:results_contact_edges}
    \vspace{-2mm}
\end{figure}

We adopt the depth completion pipeline proposed by \citet{zhang2018deepdepth} with a few critical modifications to address the unique challenges presented by transparent objects. 
First, instead of only filling in the missing depth regions, we train an additional network to predict a pixel-wise mask for transparent surfaces and use it to remove unreliable depth measurements from the depth camera, see Fig. \ref{fig:sensor_errors}.
Second, instead of predicting only occlusion edges (discontinuities in depth), we propose to predict both occlusion and contact edges (boundaries of objects in contact with other surfaces) so that the network can distinguish different type of edges and predict more accurate depth discontinuity boundaries, which is critical for the global optimization step, see Fig. \ref{fig:results_contact_edges}. 
Fig. \ref{fig:method} shows an overview of our approach. The following paragraphs provide details on each module.

\myparagraph{Transparent object segmentation.}
Due to the reflective and refractive nature of transparent objects, they cause erroneous readings in commodity RGB-D sensors.
\figref{fig:sensor_errors} explains 2 types of errors. Type I error refers to missing depth, commonly caused by specular highlights. Type II error occurs when the light is refracted through the transparent material, only to reflect back from the surface behind the object. This causes the sensor to report the depth of surfaces behind the object instead of the object itself. These inaccurate non-zero depth estimates are difficult to detect using standard depth completion, which would only propagate the inaccurate depth and result in corrupted reconstructions.
To address this issue, we predict the pixel-wise masks of transparent objects using a Deeplabv3+ \cite{chen2018encoder} with a DRN-D-54 backbone \cite{Yu2017} to remove all depth pixels corresponding to transparent surfaces.

\myparagraph{Surface normal estimation.}
This module predicts pixel-wise surface normals for the input RGB color image using Deeplabv3+ with DRN-D-54. The last convolutional layer is modified to have 3 output classes. To ensure that estimated normals are unit vectors, the output is L2 normalized. 

\myparagraph{Boundary detection.}
This module labels each pixel of the input color image as one of three classes: (a) Non-Edge, (b) Occlusion Boundary (depth discontinuity) (c) Contact Edges (points of contact between 2 objects). Contact edges, while not directly used by the optimization step, is very important because it helps the network better distinguish between different types of edges observed in color images, and therefore results in more accurate predictions of depth discontinuity boundaries.
This significantly decreases chances of the model predicting a boundary around an entire object, which would prevent the global optimization step from solving back its depth using predicted surface normals.   
We use the same Deeplabv3+ model with a DRN-D-54 backbone. Since the pixel ratio of boundaries to background is low, we use a weighted cross-entropy loss with boundary pixels weighing 5x more than background pixels. 

\myparagraph{Global optimization for depth.}
Using the depth image (with all pixels corresponding to transparent surface removed) and predictions of surface normals and occlusion and contact edges, ClearGrasp reconstructs the 3D surfaces of transparent objects (missing depth region) via the global optimization algorithm proposed by \citet{zhang2018deepdepth}. 
The optimization algorithm fills in the removed depth using the predicted normals to guide the shape of the reconstruction, while observing the depth discontinuities indicated by the occlusion boundaries.
It solves a system of equations with the goal of minimizing the weighted sum of squared errors of three terms:
$E = \lambda _{D} E_{D} + \lambda _{S} E_{S} + \lambda _{N} E_{N} B $, where $E_{D}$ measures the distance between the estimated depth and the observed raw depth, $E_{S}$ measures difference between the depths of neighboring pixels and $E_{N}$ measures the consistency between estimated depth and predicted surface normal. $B$ down-weights the normal terms based on the predicted probability that a pixel is on an occlusion boundary. In our experiments: $\lambda _{D}=1000$, $\lambda _{S}=0.001$ and $\lambda _{N}=1.0$.

\begin{figure}[ht]
    \centering
    \includegraphics[width=1.0\linewidth]{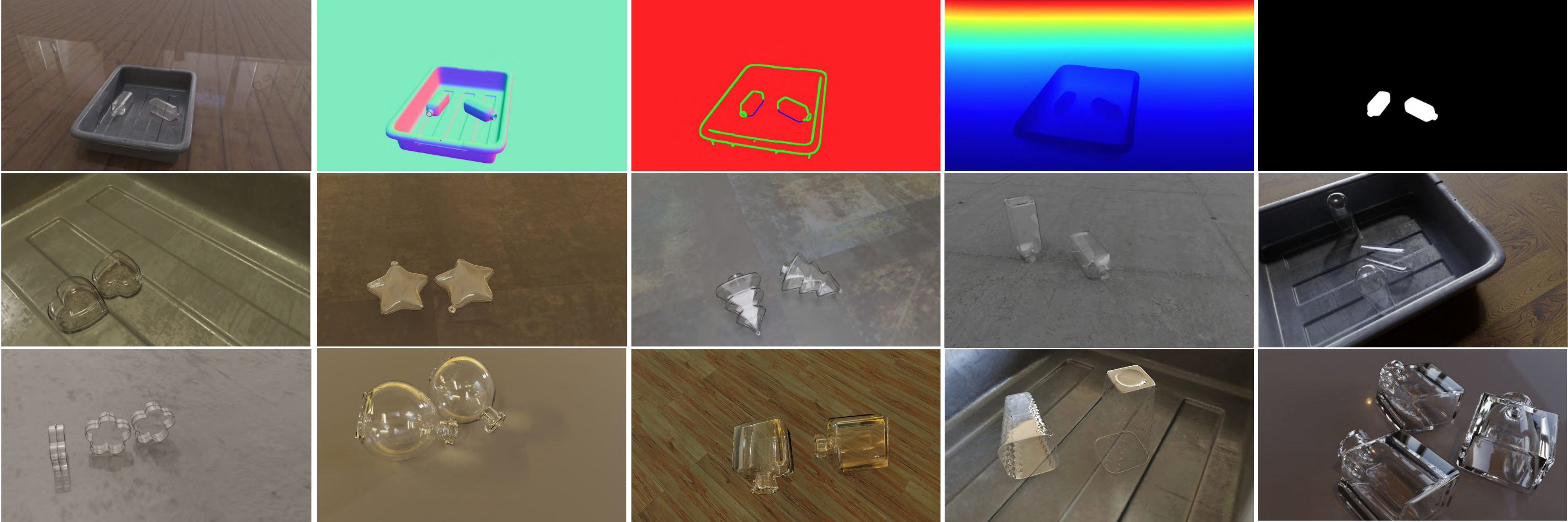}
    \caption{ \label{fig:train_data}　\textbf{Synthetic data.} Top row is the rendered image and its groundtruth (surface normal, boundary, depth and mask). Bottom two rows are rendering of different objects.}
    \vspace{1mm}
    
    \centering
    \includegraphics[width=1.0\linewidth]{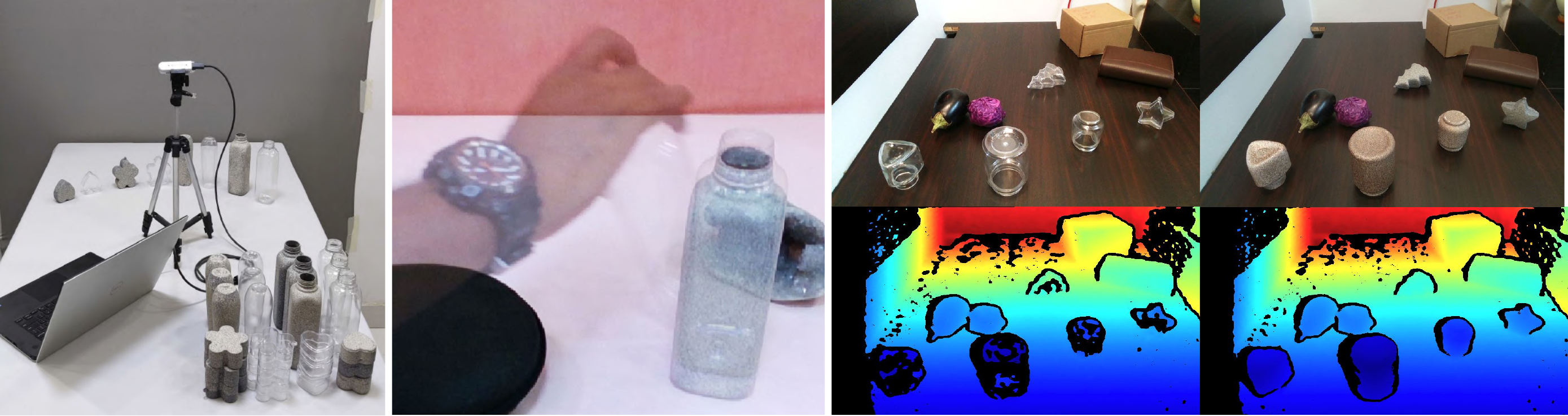}
    \caption{\label{fig:real_data}　\textbf{Real-world benchmark.} From left to right: the data capturing process, screenshot of GUI showing the process of replacing transparent bottle with opaque bottle, RGB-D images of transparent objects, RGB-D images of replaced spray-painted objects.
    }
    \vspace{-4mm}
\end{figure}

\subsection{Synthetic Training data generation}
\label{sec:data-syn}
We selected Synthesis AI's platform to generate our synthetic data, using Blender's physics engine\cite{blender}, as well as the physically-based, ray-tracing Blender Cycles \cite{Blendercycles} rendering engine.  We selected this because it is highly configurable, and is able to simulate important effects for transparent objects like refraction and reflection through multiple surfaces, as well as soft shadows.

The dataset consists of 9 CAD models modeled after real-world transparent plastic objects, in which we hold out 4 of the objects during training to test the algorithm generalization ability. Additionally, one gray tote box is used as an background object. We employed 33 HDRI lighting environments and 65 textures for the ground plane underneath the transparent objects.  
Camera intrinsics were set to that of the Intel RealSense D415 camera.  
To generate each scene, between 1 and 5 CAD model objects were created above the plane surface, with or without a gray tote box and the CAD model objects were “dropped” so they would come to rest according to physics. Then, a random selection of HDRI lighting environments and ground plane surface textures would be applied to each scene as well.  

For each scene, the ground truth data includes: (1) monocular RGB render, (2) aligned depth in meters, (3) semantic segmentation of all transparent objects, (4) pose of the camera (5) pose of each CAD object, and (6) surface normals of the scene.
Fig. \ref{fig:train_data} shows example rendered images and their corresponding ground truth geometry.
The final training dataset consists of over 13,000 images of 3 objects each and 5000 images each of another 2 objects. 100 images of each were kept aside as a validation set. For the test set, we rendered around 100 images of each of the 4 testing objects.

\subsection{Real-World Benchmark}
\label{sec:data-real}
To test the ability of our model to generalize to real-world images, we create a dataset of real-world transparent objects.
The setup consists of a photography background cloth or wooden laminate spread across a flat surface kept against a wall. Five unique wooden laminates and five different background cloths were used. 
The scene was lit with ambient lighting to avoid sharp caustics.
The camera was mounted on a tripod at a distance of 40-100cm from the objects.

To capture the depth of transparent objects, we separated the objects into 2 equal sets and spray painted one set with a rough stone texture, which gives much better depth than a flat color. 
A GUI app was developed that could overlay 2 frames read from the camera, as shown in Fig. \ref{fig:real_data}. First the transparent objects were placed in the scene along with various random opaque objects like cardboard boxes, decorative mantelpieces and fruits. After capturing and freezing that frame, each object was replaced with an identical spray-painted instance. Subsequent frames would be overlaid on the frozen frame so that the overlap between the spray painted objects and the transparent objects they were replacing could be observed. With high resolution images, sub-millimeter accuracy can be achieved in the positioning of the objects.

The validation dataset consists of 173 images of 5 known objects used in synthetic training data. The testing set consist of 113 images of 5 novel objects, including 3 new glass objects not present in the synthetic dataset. Each image contains 1-6 objects, with an average of 2 objects per image.

\begin{figure*}[t]
    \centering
    \vspace{-2mm}
    \includegraphics[width=1.0\linewidth]{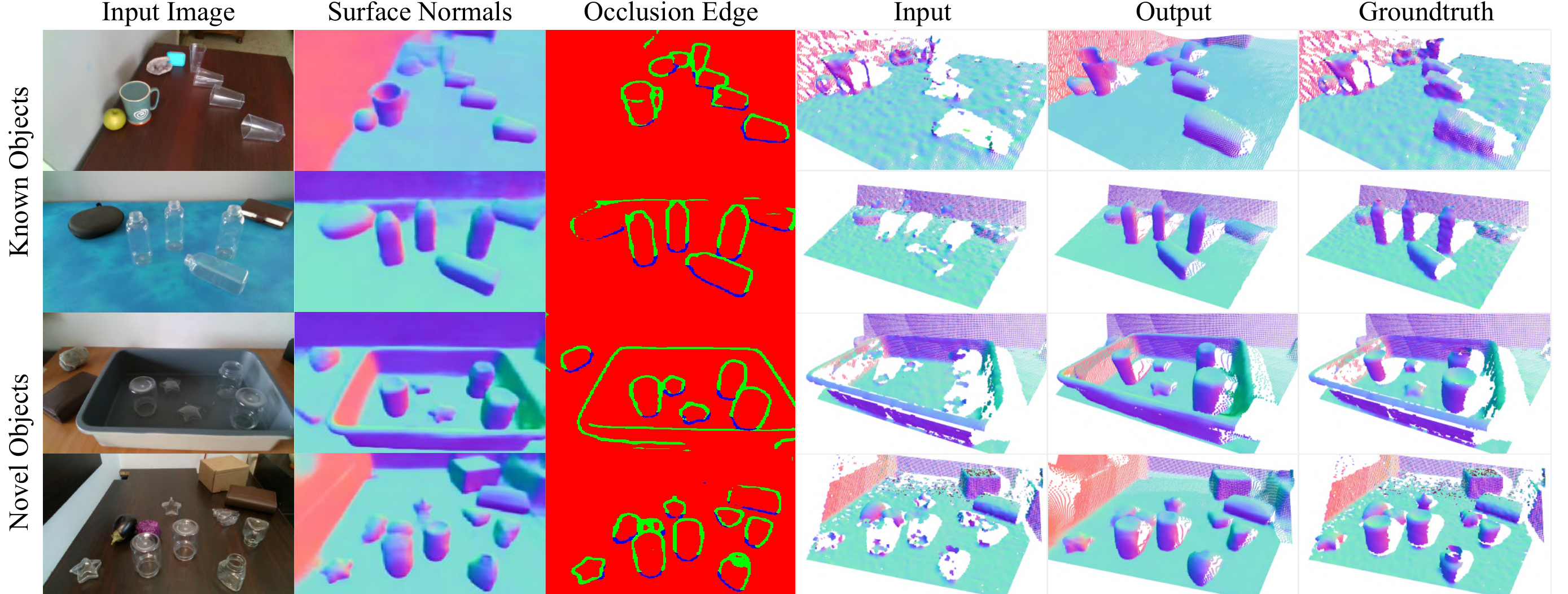}\vspace{-2mm}
    \caption{\textbf{Qualitative results} on real-world benchmark with known objects (rows 1-2) and novel objects (rows 3-4). More results can be found in the supplementary material website.}
    \label{fig:results_collage}
    \vspace{-4mm}
\end{figure*}

\subsection{Grasp planning}
\label{sec:grasping}
By integrating ClearGrasp into a robotic picking system, we can investigate its benefits for downstream manipulation tasks. We adapted a state-of-the-art grasping algorithm for our experiment \cite{zeng2018robotic,zeng2018learning,zeng2019tossingbot}, which consists of a convolutional network that predicts the probability of picking success for a scripted grasping primitive across a dense pixel-wise sampling of end effector locations and orientations across the completed depth images from ClearGrasp. Specifically, it uses an 18-layer fully convolutional residual network \cite{he2016deep} with dilated convolutions \cite{Yu2017} and ReLU activations \cite{nair2010rectified}, interleaved with 2 layers of max pooling, 2 layers of spatial bilinear $2x$ upsampling. 
The network takes as input a 4 channel image -- the surface normal map (3 channels) concatenated channel-wise with the completed depth image (1 channel) inferred from ClearGrasp -- and outputs a probability map with the same size and resolution as that of the input image. The picking system assumes that the 3D camera is calibrated with respect to robot coordinates using the calibration procedure in \cite{zeng2018learning} -- hence each pixel in the depth image maps to a 3D location. The robot executes a top-down parallel-jaw grasp or suction where the tip of the end effector is centered at the 3D location of the pixel with the highest predicted probability from the network.

For our experiments with parallel-jaw grasping, as in \cite{zeng2018robotic} we account for different grasping angles by constructing top-down orthographic heightmaps from ClearGrasp depth images, rotating the input heightmaps by 16 orientations (multiples of $22.5^\circ$), and feeding each heightmap through the network for a total of 16 forward passes. The pixel and the corresponding rotation with the highest predicted probability among all 16 maps determines the respective grasping angle. The network is trained end-to-end using the binary cross-entropy error from predictions of grasp success against the binary ground truth success labels. We pass gradients only through the single pixel on which the grasping primitive was executed. Since each pixel-wise prediction shares convolutional features for all grasping locations and orientations, the network is sample-efficient and trains within a few hundred trial-and-errors.

\section{Evaluation}
We evaluate ClearGrasp's ability to estimate transparent object geometry on both synthetic and real-world benchmarks, then apply it to a real-world robotic picking system.  

\begin{table*}[t]
    \centering
    \caption{\textbf{Generalization.} ClearGrasp generalizes to both real images and novel transparent objects unseen in training.  \label{table:results_depth}} \vspace{-2mm}
    \setlength\tabcolsep{5.0pt}
     
    \begin{tabular}{ll|cccccc|ccccc|cc}
        \toprule
        \multicolumn{2}{c}{Testset} & \multicolumn{6}{|c|}{Depth Estimation} & \multicolumn{5}{c}{Surface Normal Estimation} & \multicolumn{2}{|c}{Mask}  \\
        \midrule
          Type & Object & RMSE$\downarrow$ & REL$\downarrow$ & MAE$\downarrow$ & $\delta_{1.05}\uparrow$ & $\delta_{1.10}\uparrow$ & $\delta_{ 1.25}\uparrow$ & mean$\downarrow$ & med.$\downarrow$ & $11.25\degree\uparrow$ & $22.5\degree\uparrow$ & $30\degree\uparrow$ & IoU & TP \\
        \midrule
        Synthetic & Known & 0.044 & 0.047 & 0.033 & 71.23 & 92.60 & 98.24 & 15.64 & 10.62 & 53.71 & 78.28 & 85.83 & 0.93 & 95.90 \\
        Synthetic & Novel & 0.040 & 0.071 & 0.035 & 42.95 & 80.04 & 98.10 & 25.32 & 20.53 & 24.04 & 55.88 & 69.73 & 0.94 & 97.58 \\
        \midrule
        Real    & Known & 0.039 & 0.053 & 0.029 & 70.23 & 86.98 & 97.25 & 21.93 & 18.72 & 32.82 & 64.39 & 76.05 & 0.63 & 96.30 \\
        Real    & Novel & 0.028 & 0.040 & 0.022 & 79.18 & 92.46 & 98.19 & 22.29 & 18.09 & 31.63 & 63.44 & 76.06 & 0.58 & 96.95 \\
        \bottomrule 
    \end{tabular}
   \vspace{-2mm}
   
\end{table*}

\textbf{Datasets} used to train and test our algorithm: 
\begin{itemize}
    \item Syn-train: Synthetic training set with 5 objects as described in Sec. \ref{sec:data-syn}.
    \item Syn-known: Synthetic validation set for training objects.
    \item Syn-novel: Synthetic test set of 4 novel objects.
    
    \item MP+SN: Out-of-domain real-world RGB-D datasets of indoor scenes that do not contain transparent objects' depth (Matterport3D \cite{Matterport3D} and ScanNet \cite{dai2017scannet}).
    \item Real-known: Real-world test set for all 5 of the training objects. Sec. \ref{sec:data-real} describes the capturing procedure.
    \item Real-novel: Real world test set of 5 novel objects, including 3 not present in synthetic data.
\end{itemize}

\myparagraph{Metrics:}
For surface normal estimations, we calculate the mean and median errors (in degrees) and the percentages of pixels with estimated normals less than thresholds of 11.25, 22.5, and 30 degrees.
For depth estimation, we use metrics standard among previous works \cite{corr2014EigenPF}: the Root Mean Squared Error in meters (RMSE), the median error relative to the depth (Rel) and percentages of pixels with predicted depths falling within an interval ([$ \delta = |predicted - true| / true$], where $\delta$ is $1.05$, $1.10$ or $1.25$). Depth is evaluated by resizing the images and ground truth to 144x256p resolution.
For mask prediction, we use pixel-wise intersection over union for evaluation as well as true positive rate. 
Unless specified, metrics are calculated on the real-known dataset only over the pixels belonging to transparent objects.

\myparagraph{Generalization: real-world images.}
Table \ref{table:results_depth} also shows results on an experiment to test the cross-domain performance of our models. Despite never being trained on real-world transparent objects, we find our models are able to adapt well to the real-world domain achieving very similar RMSE and Rel scores on known objects across domains. However, the surface normal prediction accuracy decreases on real images. We observe large errors in surface normal estimations when transparent object occlude novel opaque objects. 
Surprisingly, the metrics of Real-novel objects are better than Syn-novel. We attribute this to the 3 new glass objects used in real-world images which show more evident refraction characteristics due to their thicker material as compared to the thin plastic material of all other objects.

\myparagraph{Generalization: novel object shapes.}
We inspect the ability of our algorithm to generalize to previously unseen object shapes. Table \ref{table:results_depth} shows the results of depth estimation on novel objects, conducted on both synthetic data and real-world data. We see that it is able to generalize remarkably well in both cases, achieving better results than on the known objects. This is likely due to the smaller size of novel objects, which cause a relatively smaller error in depth reconstruction.

\begin{table}[ht]
\setlength\tabcolsep{2.pt}
\centering
\caption{\textbf{Baseline comparisons and ablation study.}}
\vspace{-2mm}
\begin{tabular}{l|cccccc}
\toprule
& RMSE$\downarrow$ & REL$\downarrow$ & MAE$\downarrow$  & $\delta_{1.05}\uparrow$ & $\delta_{1.10}\uparrow$ & $\delta_{1.25}\uparrow$ \\
\midrule
DenseDepth \cite{Alhashim2018densedepth} & 0.270 & 0.428 & 0.259 & 18.67 & 34.34 & 58.29 \\
DeepCompletion \cite{zhang2018deepdepth} & 0.054 & 0.081 & 0.045 & 44.53 & 69.71 & 95.77 \\
\midrule
- Mask          & 0.054 & 0.080 & 0.044 & 44.46 & 69.73 & 96.06 \\
- Contact Edge  & 0.061 & 0.096 & 0.054 & 36.64 & 65.11 & 92.38 \\
- Edge Weights  & 0.049 & 0.075 & 0.042 & 51.77 & 73.70 & 95.59 \\
Full            & \textbf{0.038 }& \textbf{0.048} & \textbf{0.027} & \textbf{72.94} & \textbf{87.88} &\textbf{97.17} \\
\bottomrule
\end{tabular}
\vspace{-5mm}
\label{table:ablation}
\end{table}

\myparagraph{Comparison with Monocular Depth Estimation.}
We compare our approach with DenseDepth \cite{Alhashim2018densedepth}, a monocular depth estimation method that has state-of-the-art performance. DenseDepth uses a deep neural network to directly predict the depth value from the color image. We train DenseDepth with the same training data as our approach.
The results in \tabref{table:results_depth} show that our model outperforms the monocular depth estimation methods by a large factor.

\myparagraph{Effect of mask prediction.}
We test the effectiveness of cleaning the input depth by removing all pixels belonging to transparent objects, as shown in Table \ref{table:ablation}. By not removing the initial noisy  depth values, we notice a significant increase in the final depth estimation error. 
Table \ref{table:results_depth} reports the accuracy of the mask prediction in both intersection over union and true positive rate. In our approach, having a high true positive rate ($>$ 95\%) is critical for removing all the incorrect initial depth values.  

\myparagraph{Effect of contact edges and edge weights}
Table \ref{table:ablation} shows the effect of using a weighted loss function and the impact of adding the additional class of contact edges to our occlusion boundary estimation model. Both of these methods contribute significant improvement in depth completion results.

\begin{table}[t]
    \caption{\textbf{Training data.} Normal estimation performance under different training procedures: with/without out-of-domain realworld data (MP+SN) and in-domain synthetic data (Syn). 
    \label{table:real_pretraining}}
    \vspace{-2mm}
    {\centering
    \setlength\tabcolsep{5.8 pt}
    \begin{tabular}{cc|ccccc}
        \toprule
        Pretrain & Train & Mean$\downarrow$ & Median$\downarrow$ & $11.25\uparrow$ & $22.5\uparrow$ & $30\uparrow$ \\
        \midrule
        MP+SN &  -  & 43.92 & 45.31 & 9.51 & 22.69 & 32.03 \\
        -     & Syn & \textbf{21.59} & 24.74 & 24.74 & 55.97 & 70.40 \\
        MP+SN & Syn & 21.93 & \textbf{18.72} &\textbf{32.82} & \textbf{64.39} &\textbf{76.05} \\
        \bottomrule 
    \end{tabular}
    }
    \vspace{1mm}
    \includegraphics[width=1.0\linewidth]{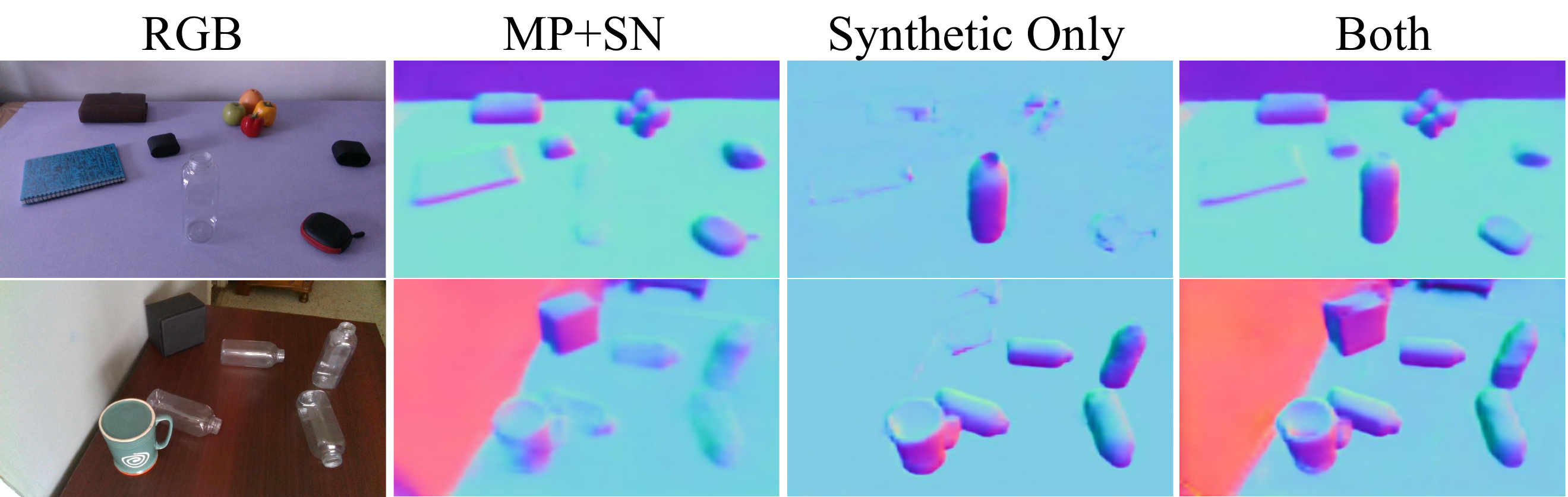}     
    \vspace{-6mm}
\end{table}

\myparagraph{Effect of training data.}
Our main training dataset consists of synthetic images of transparent objects. Since it is expensive to capture real-world data with accurate geometry ground truth for transparent objects, we propose to mix in typical real-world RGB-D indoor scene images in training to reduce the domain gap. 
Table \ref{table:real_pretraining} shows the model performance under different training procedure: with/without pre-training on out-of-domain real-world data (80k images from the Scannet and Matterport datasets) and with/without in-domain synthetic data fine-tuning.  Fig. \ref{fig:results_collage} additionally shows the qualitative results of surface normal estimation for all the above cases. 
We see that a model trained on out-of-domain real-world data is not able to pick up transparent objects. However, pre-training with such data improves results, especially for real-world test sets.

\myparagraph{Robot manipulation.}
We also incorporate ClearGrasp as part of a real-world robotic picking system to observe how it influences the overall grasping performance of transparent objects. In this experiment, a pile of 3 to 5 transparent objects are presented on a table within the robot's workspace, of which RGB-D images are captured using a calibrated RealSense camera. The goal of the robot is to pick up objects from the table using a state-of-the-art grasping algorithm described in Sec. \ref{sec:grasping}. Fig. \ref{fig:robot_grasping} shows the setup. 
We test the algorithm using two end-effectors: suction and a parallel-jaw gripper. For each end-effector type, with and without ClearGrasp, we train a grasping algorithm using 500 trial and error grasping attempts, then test it with 50 attempts. We compute the average grasping success rate $=\frac{\mathrm{\#~successful~picks}}{\mathrm{\#~picking~attempts}}$  \cite{zeng2018learning} as the evaluation metric.
With both end-effectors, we observe that ClearGrasp significantly improves the grasping success rate of transparent objects: it improves the grasping success from  64\%  to 86\% for suction, and 12\%  to 72\% for parallel-jaw grasping.

\section{Conclusion and Future Work} 
\label{sec:conclusion}
We present ClearGrasp, an algorithm that leverages deep learning with synthetic training data and multiple sensor modalities (color and depth) to infer accurate 3D geometry of transparent objects for manipulation.
However, the proposed system is still far from perfect. Possible future directions may include: explicitly leveraging lighting information during the inference step to improve the algorithm's accuracy under different lighting conditions, improving the algorithm robustness in cluttered environments where predicting accurate occlusion and contact edges is more challenging and making the algorithm robust to sharp caustics and shadows.

\newpage
\bibliographystyle{plainnat}
\bibliography{references.bib}

\clearpage
\section*{APPENDIX}

The appendix consists of additional system details, analysis, and experimental results.

\subsection{Additional Details on Dataset}
\begin{figure}[ht]
    \centering
    \includegraphics[width=1.0\linewidth]{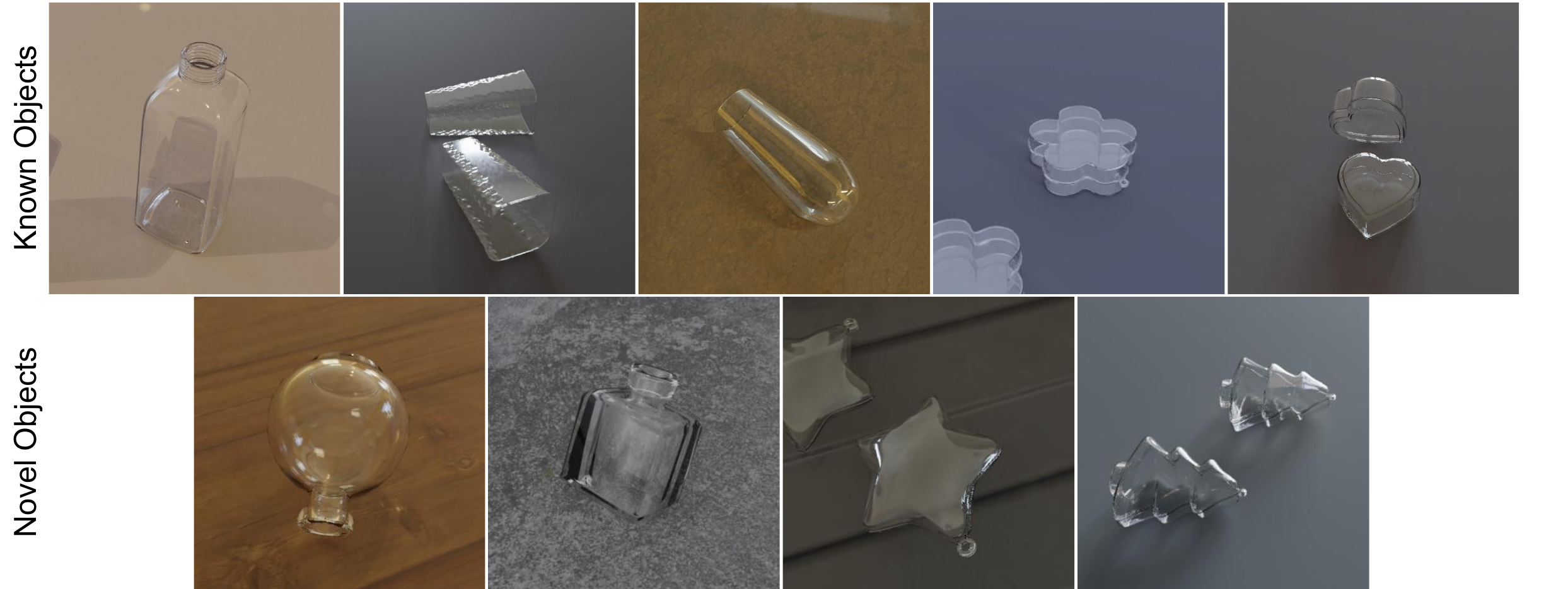}
    \caption{\textbf{Known and novel objects in Synthetic dataset.} We have 5 known objects for training and 4 novel objects for testing. The test objects are all challenging: 2 are of thick glass (a different material from our plastic known objects) and 2 are of complex shapes.}
    \label{fig:samples_datasets_syn}
    \vspace{-2mm}
\end{figure}

\begin{figure}[ht]
\vspace{-1mm}
    \centering
    \includegraphics[width=1.0\linewidth]{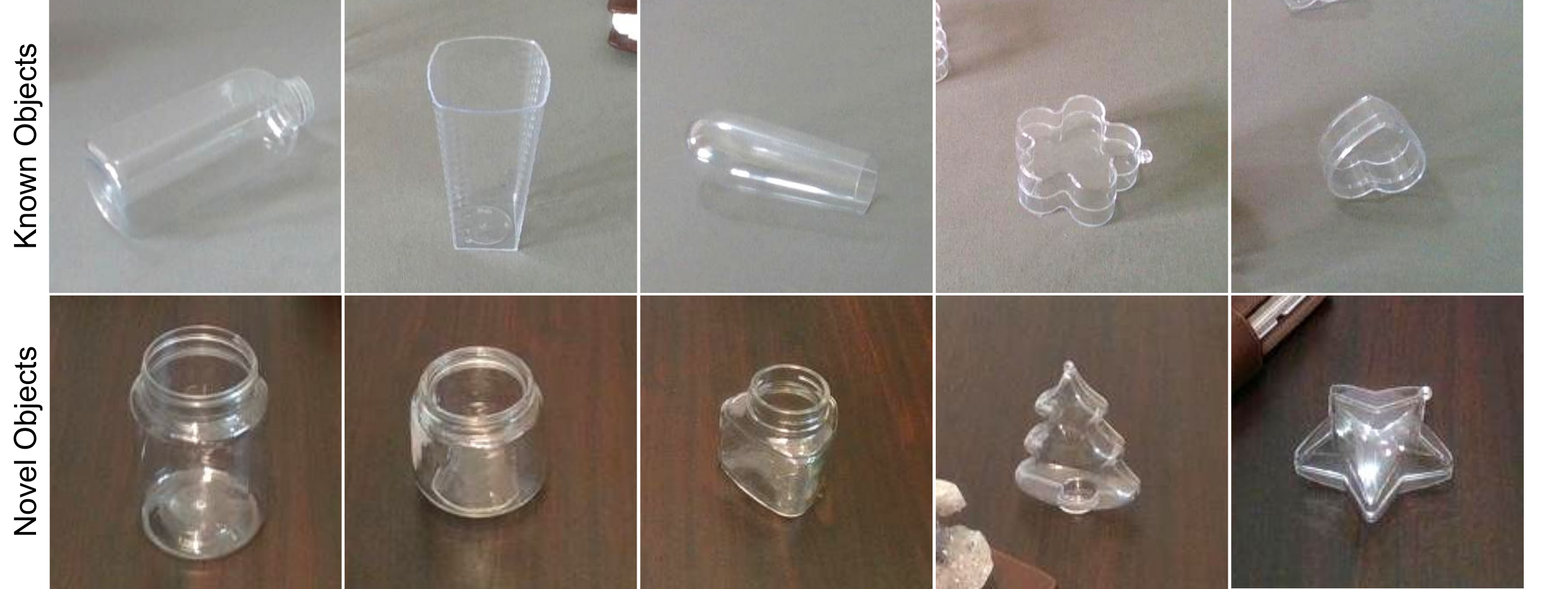}
    \caption{\textbf{Known and novel objects in Real-World benchmark dataset.} We have 5 known objects and 5 novel objects. All 5 known objects and 2 of the novel objects were used to model the synthetic objects. 2 other novel objects are made of glass instead of plastic.}
    \label{fig:samples_datasets_real}
    \vspace{-2mm}
\end{figure}

Fig. \ref{fig:samples_datasets_syn} and Fig. \ref{fig:samples_datasets_real} showcase the objects used within our synthetic and real-world datasets. In the real-world dataset, all images of known objects are taken with a RealSense D435 camera and 80\% of the images of novel objects were taken with a RealSense D415 camera instead.

\subsection{Limitations and Failure Cases}
We detail some of the failure cases of our models. Four examples are shown in Fig. \ref{fig:failure_cases}

\begin{itemize}[leftmargin=*]
\item The biggest limitation of our approach is that it is not always possible to reconstruct depth from surface normals directly \cite{zhang2018deepdepth} - if a region is completely enclosed by an occlusion boundary, its depth is left indeterminate from the rest of the scene. In Case I, we see a bottle that is partially occluded, with its contact edges not visible. In Case II, we see the mouth of the glasses cause the inner portions to be (correctly) completely enclosed by an occlusion boundary. In both cases, the depths of such regions become indeterministic and can be assigned random values.

\item Cluttered scenes are challenging. In cases where multiple transparent objects are partially or completely occluding each other, it becomes challenging to correctly predict surface normals and occlusion boundaries, which leads to errors in the output depth. Case III highlights such a scenario. Another situation which our models find challenging is when the background seen behind a transparent object is not constant - such as when a bottle is at the edge of a table or when its partially occluding an opaque object.

\item As seen in Case IV, bright directional lighting and its associated caustics cause our model to mistakenly identify shadows of transparent objects as transparent objects. Our models seems to pick up on cues like specular highlights to identify transparent objects and may be confusing the caustics on shadows with specular highlights - hence detecting the shadow as a transparent object. Since our synthetic dataset does not contain accurate caustics due to the limitations of the Cycles rendering engine, our model is particularly susceptible to this problem.
\end{itemize}

\subsection{Additional Training Details}
We make use of Deeplabv3+ with a DRN-D-54 backbone \cite{deeplabv3plus} in Pytorch for all 3 of our neural networks - surface normal estimation, occlusion boundary prediction and segmentation of transparent surfaces. For all 3 networks, we start with a model pre-trained for semantic segmentation on the COCO dataset and use the same hyperparameters: SGD Optimizer with constant learning late of 1e-6, momentum 0.9 and weight decay 5e-4. We used a GCP server with 8x Nvidia V100 GPUs enabling a batch size up to 128 at an input image size of 256x256p.

For surface normals, we initially pre-train our model on the Matterport3D (MP) and Scannet (SN) datasets by selecting a random subset of approximately 40k images from each, for a total dataset of 80k images. When training on transparent objects, we include a new random subset of 2k images from MP and 2k from SN each epoch. Our synthetic dataset contains only a flat plane and up to 5 transparent objects, lacking any other surfaces like walls and random opaque objects. Injecting MP+SN images every epoch allows the model to retain knowledge of the previous domain and predict more accurate normals for surfaces like walls.
To train the model more quickly on the different task of surface normal estimation, we adopted a staggered training approach: First, we trained a small subset of ~100 images at a reduced resolution of 128x128p. Second, we took an early checkpoint before the model starts to stabilize and train on a larger subset of our data. This step was repeated on subsequently larger subsets. Third, we repeat the procedure with the larger image size of 256x256p taking the checkpoint from the previous step.

To make the models more robust, the following data augmentations were utilized from the imgaug \cite{imgaug} library:
\begin{itemize}[leftmargin=*]
\item Flip Up-Down
\item Flip Left-Right (not used for surface normals)
\item Rotate 90 degrees (not used for surface normals)
\item Color Space Augmentations: Add (RGB), Multiply \\
(RGB), Add Hue, Add Saturation, Contrast Normalization, Grayspace
\item Blur: Motion Blur or Gaussian Blue
\item Noise: Add Element-wise, Multiply Element-wise, Additive Gaussian Noise, Additive Laplacian Noise, Dropout
\item Large Patches: 
    \begin{itemize}
         \item Channel-Wise Coarse Dropout up to 1/4th the size of the image. This makes the model more robust to varying backgrounds seen behind a transparent object.
         \item Bright White Patches: We use a Simplex Noise blended with a white image to generate random white patches which are overlaid with transparency on our images. Since our synthetic images do not contain significant caustics, this augmentation attempts to make the model more robust to bright patches of light due to caustics or directional lights.
    \end{itemize}
\end{itemize}

Using this data augmentation strategy, we noticed a significant improvement in scenes with a patterned background cloth, bright caustics or directional lights and cases where the background behind a transparent object varied (like when it partially occludes an opaque object).

\subsection{Experiment on Network Architectures}
During our trials, we noticed that surface normal estimation was better on bottles that were kept further away from the camera. This led us to hypothesize that a larger receptive field might be helpful for transparent objects. \tabref{table:receptive_field} shows the results of experimenting with different models and input image sizes. We try Deeplabv3+ with 2 different backbones: Resnet-101 and DRN-54 (Dilated Residual Network). We also experiment with different input image sizes. The results indicate that a smaller input image size, which effectively increases receptive field width, performs better. Further, replacing Resnet with DRN, which increases the receptive field of the network \cite{Yu2017}, improves results even more - hence validating our hypothesis.

\begin{table}[ht]
    \centering
    \setlength\tabcolsep{3px}
    \begin{tabular}{l|c|ccccc}
        \toprule
        Backbone & Input Size & Mean$\downarrow$ & Median$\downarrow$ & $11.25\uparrow$ & $22.5\uparrow$ & $30\uparrow$ \\
        \midrule
        Resnet101  & 512 & 34.9 & 32.8 & 16.0 & 42.6 & 56.8 \\
        Resnet101  & 256 & 25.3 & 22.2 & 25.7 & 56.9 & 70.0 \\
        DRN-54 [ours] & 256 & \textbf{22.5 }& \textbf{19.4} & \textbf{28.6} & \textbf{61.5} & \textbf{ 75.5 }\\
        \bottomrule 
    \end{tabular}
    \caption{Network architectures for surface normal.}
    \label{table:receptive_field}
\end{table}

\subsection{Additional Qualitative Results}
\begin{figure*}[pht]
  \centering
  \includegraphics[height=0.45\textheight]{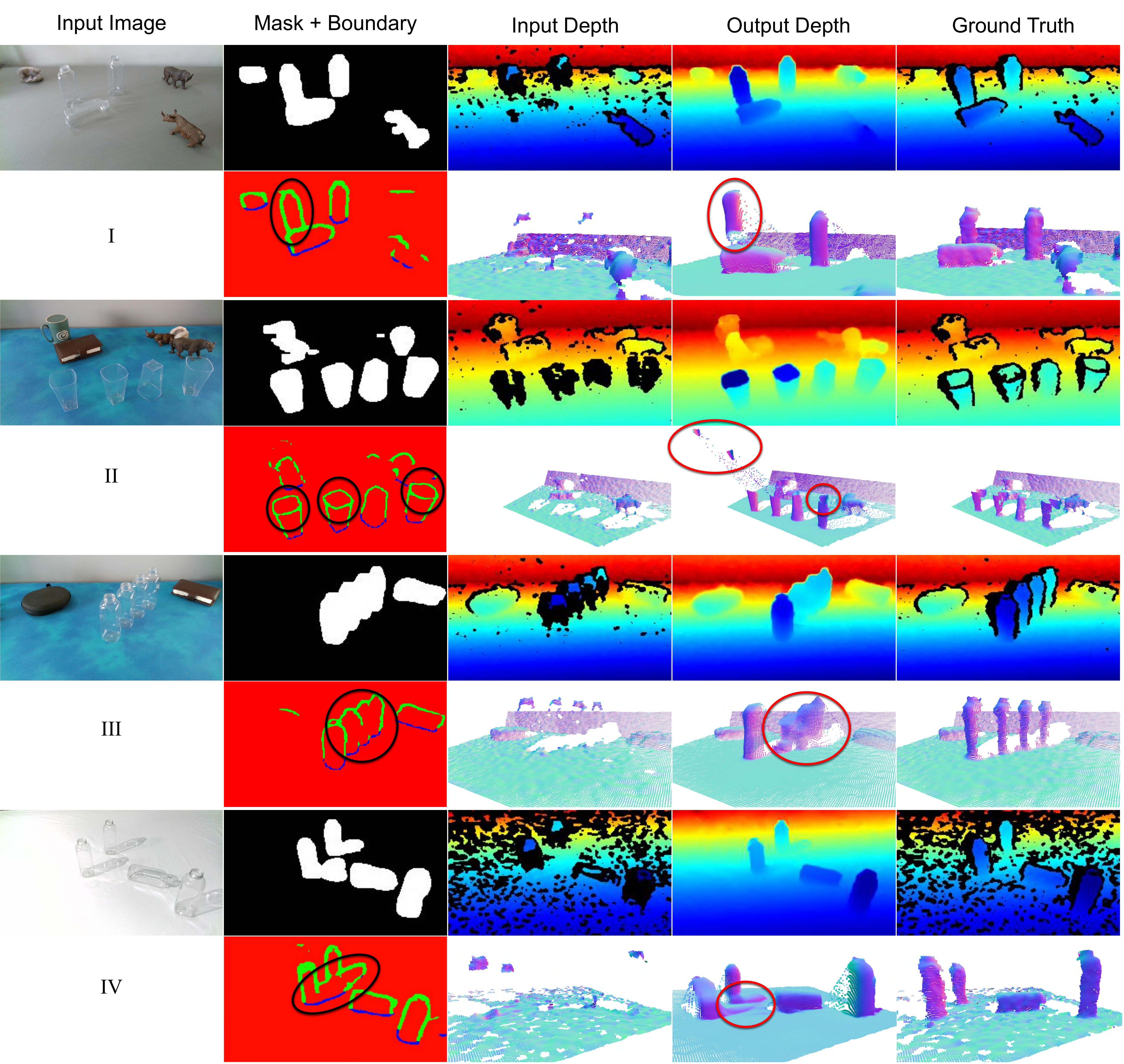}
  \caption{\textbf{Failure Cases.} Most of the errors in output depth (highlighted in red) are due to the errors in occlusion boundary prediction (highlighted in black) - either erroneous outputs or surfaces with no contact edges due to occlusion.}
  \label{fig:failure_cases}
  
  \vspace*{\floatsep}

  \includegraphics[height=0.45\textheight]{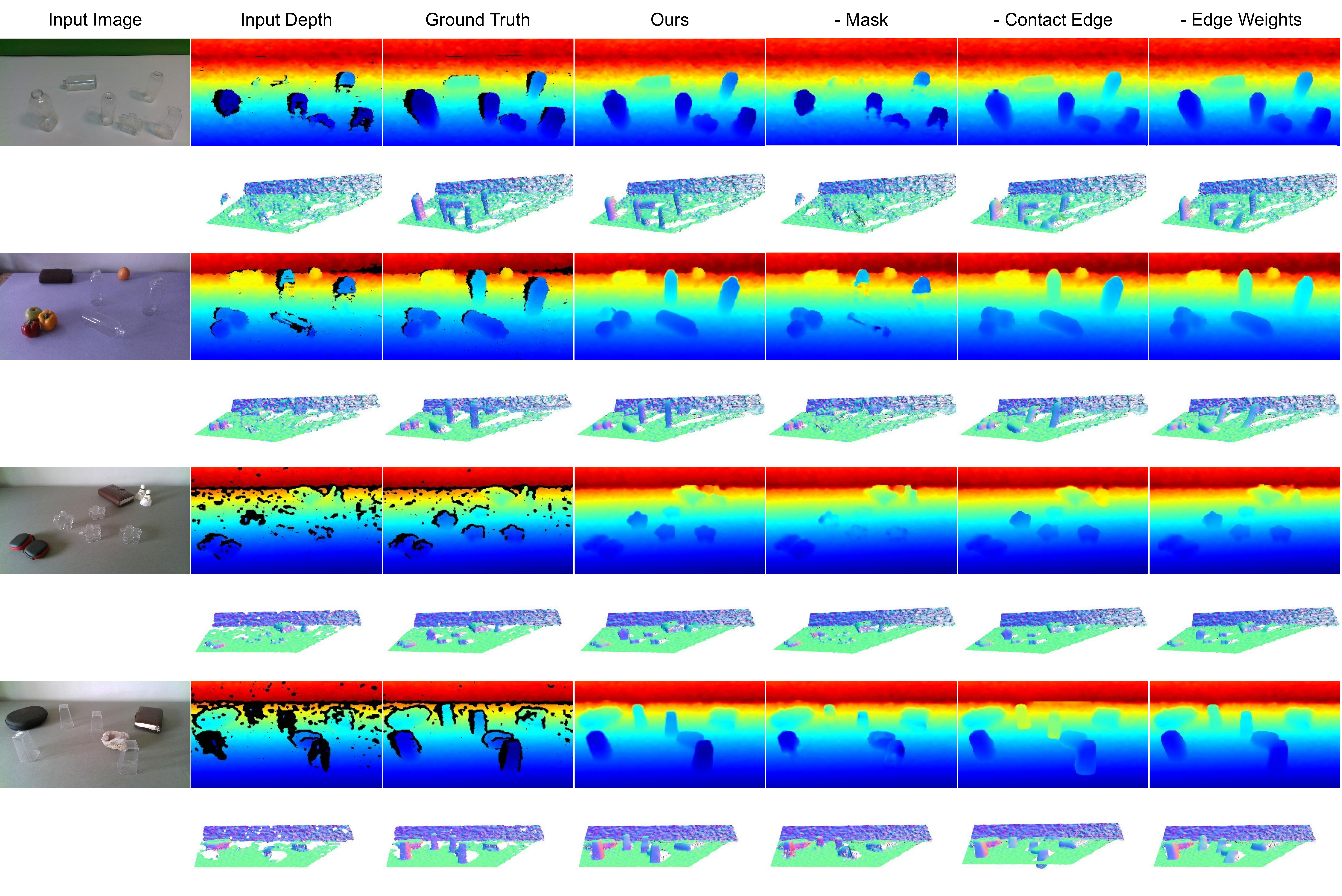}
  \caption{\textbf{Qualitative results - Ablation Study}}
  \label{fig:ablation_study}
\end{figure*}

\begin{figure*}[ph]
    \centering
    \vspace{-5mm}
    \includegraphics[width=\textwidth]{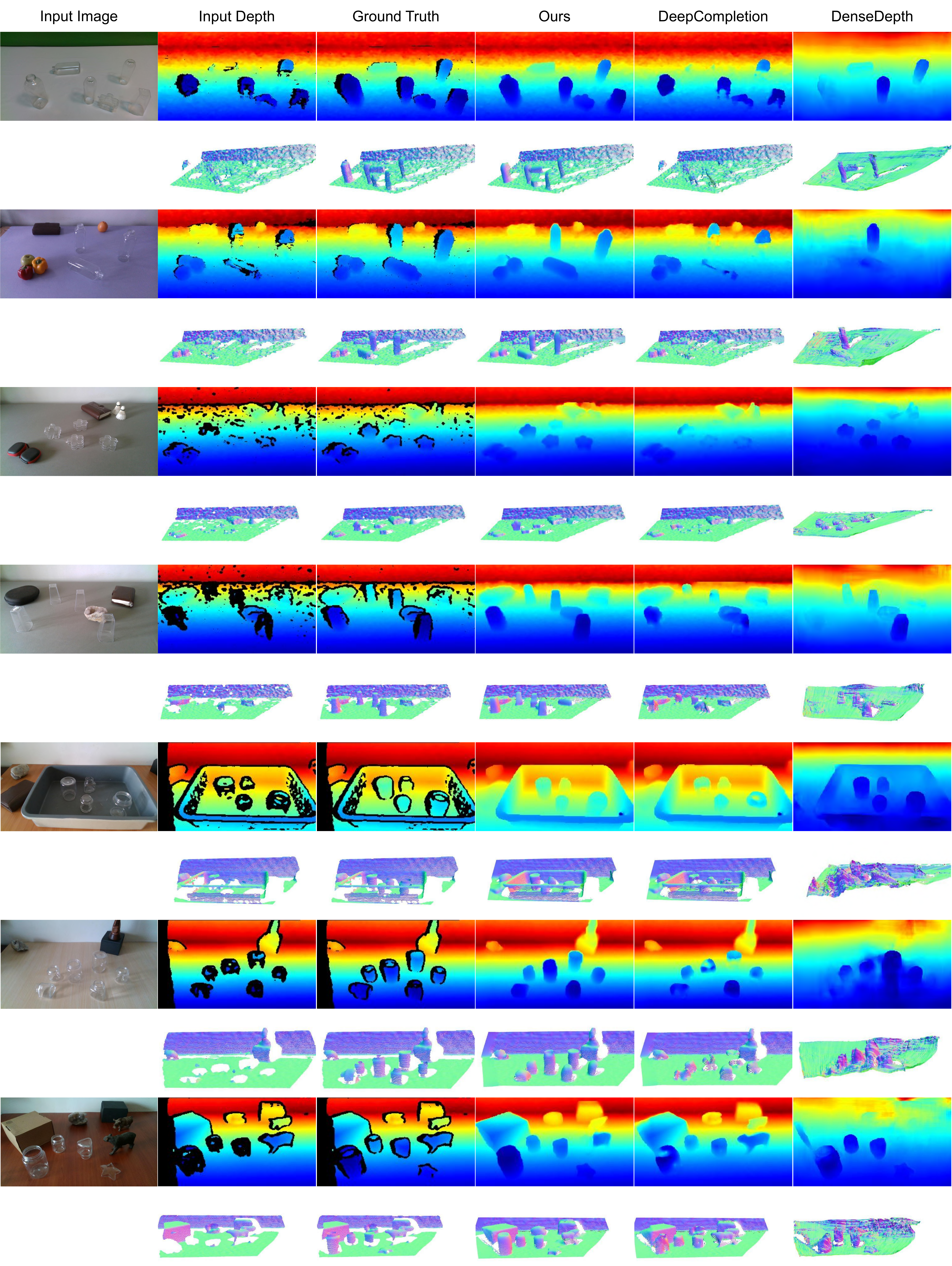}
    \vspace{-4mm}
    \caption{\textbf{Qualitative results - Comparison with baselines}}
    \label{fig:comparison_baselines}
\end{figure*}

Finally, we present some qualitative results on our ablation study and comparison with baselines (Ref to Table \ref{table:ablation}). 
Fig. \ref{fig:ablation_study} shows the performance of our approach a) without masks, b) without contact edges and c) without weighted loss terms for the contact edges.
Fig. \ref{fig:comparison_baselines} shows the qualitative results of our method in comparison with DeepCompletion and DenseDepth.
For more qualitative results please visit our \href{https://sites.google.com/view/cleargrasp}{website}. 

\end{document}